\documentclass[conference]{IEEEtran}
\usepackage{graphicx} 
\usepackage{epsfig} 
\usepackage[Symbol]{upgreek}
\usepackage{amsmath} 
\usepackage{cite}
\usepackage{lipsum}
\usepackage[font=small,skip=2pt]{caption}
\usepackage[utf8]{inputenc}
\usepackage{tikz,graphicx} 
\usepackage{pdfpages}
\usetikzlibrary{positioning}
\usetikzlibrary{decorations.markings}
\tikzstyle directed=[postaction={decorate,decoration}]
\usetikzlibrary{shapes,arrows,chains} 								
\usetikzlibrary[calc]	
\newcommand{\norm}[1]{\bigl\lVert#1\bigr\rVert}
\newcommand{\abs}[1]{\bigl\lvert#1\bigr\rvert}
\DeclareMathOperator{\mindot}{min.}
\begin{document}

\title{A Reactive and Efficient Walking Pattern Generator for Robust Bipedal Locomotion}

\author{\IEEEauthorblockN{Fatemeh Nazemi\IEEEauthorrefmark{1},
Aghil Yousefi-koma\IEEEauthorrefmark{1}\IEEEauthorrefmark{2}, Farzad A.shirazi\IEEEauthorrefmark{3}, and
Majid Khadiv\IEEEauthorrefmark{4}}
\IEEEauthorblockA{\IEEEauthorrefmark{1}Center of Advanced Systems and Technologies (CAST), Department of Mechanical Engineering,\\ College of Engineering, University of Tehran, Tehran, Iran.}
\IEEEauthorblockA{\IEEEauthorrefmark{3}Department of Mechanical Engineering, College of Engineering, University of Tehran, Tehran, Iran.}
\IEEEauthorblockA{\IEEEauthorrefmark{4}Department of Mechanical Engineering, K. N. Toosi University of Technology, Tehran, Iran.}
\IEEEauthorblockA{\IEEEauthorrefmark{2}Corresponding author: aykoma@ut.ac.ir}}

\maketitle

\begin{abstract}
Available possibilities to prevent a biped robot from falling down in the presence of severe disturbances are mainly Center of Pressure (CoP) modulation, step location and timing adjustment, and angular momentum regulation. In this paper, we aim at designing a walking pattern generator which employs an optimal combination of these tools to generate robust gaits. In this approach, first, the next step location and timing are decided consistent with the commanded walking velocity and based on the Divergent Component of Motion (DCM) measurement. This stage which is done by a very small-size Quadratic Program (QP) uses the Linear Inverted Pendulum Model (LIPM) dynamics to adapt the switching contact location and time. Then, consistent with the first stage, the  LIPM with flywheel dynamics is used to regenerate the DCM and angular momentum trajectories at each control cycle. This is done by modulating the CoP and Centroidal Momentum Pivot (CMP) to realize a desired DCM at the end of current step. Simulation results show the merit of this reactive approach in generating robust and dynamically consistent  walking patterns.
\end{abstract}


\section{{\Large I}NTRODUCTION}

Falling down for humanoid robots can cause severe damages and should be prevented at best by available tools in a motion controller. Reliable performance of humanoid robots in real environments and in the presence of disturbances demands a versatile and reactive motion generator. Trajectory optimization approaches have shown a great potential in generating reactive walking patterns for bipedal humanoid robots. Although Center of Pressure (CoP) modulation, step location and timing adjustment, and angular momentum regulation have been suggested and exploited as available tools for reacting against disturbances, the use of an optimal combination of all these tools is still a challenge.

Introducing the Linear Inverted Pendulum Model (LIPM) \cite{kajita20013d} and employing it inside the preview control of the Zero Moment Point (ZMP) \cite{kajita2003biped} have pushed the researches on bipedal walking to a fertile direction. In \cite{kajita2003biped}, Kajita et al. formulated bipedal walking pattern generation as a servo control problem on tracking a feasible ZMP trajectory in a receding horizon. In this approach, the jerk of the Center of Mass (CoM) can be manipulated to approach the measured ZMP to its desired value. Wieber \cite{wieber2006trajectory} modified this approach by considering inequality constraints on the ZMP rather than enforcing it to track a feasible trajectory. In this trajectory free approach based on Model Predictive Control (MPC), the ZMP/CoP can be freely manipulated inside the support polygon to make the planner more reactive. In \cite{diedam2008online,herdt2010online}, step location adaptation has been added to this setting which increases significantly the gait robustness, and slippage constraints are added to this new formulation in \cite{khadiv2017pattern}. Aftab et al. \cite{aftab2012ankle} added angular momentum regulation to this formulation and presented a walking pattern generator which employs CoP modulation, step location adjustment, and angular momentum regulation to stabilize stepping in the presence of disturbances. However, this approach lacks step timing adaptation, while it has been shown that it is essential to adapt timing for increasing robustness in gaits \cite{feng2015online}. To add step timing adjustment to this setting, Maximo et al. \cite{maximo2016mixed} formulated a Mixed-Integer Quadratic Program to automatically adapt step timing together with step location and CoP modulation, however this approach suffers from combinatorial complexity. Furthermore, angular momentum regulation has not been employed in this formulation.

Another stream of research uses the concept of (Instantaneous) Capture Point (CP) \cite{pratt2006capture} (which has been alternatively named eXtrapolated Center of Mass (XCoM) in \cite{hof2008extrapolated} or Divergent Component of Motion (DCM) in \cite{takenaka2009real}) which restricts only the unstable part of the CoM dynamics to generate feasible motions. In this context, Englsberger et al. \cite{englsberger2015three} generalized the DCM to 3D and generated 3D walking patterns on uneven terrain. They also proposed an analytical method to use step adjustment together with DCM tracking \cite{englsberger2017smooth}. In \cite{khadiv2016stepping}, a combination of DCM tracking and step adjustment has been used inside a hierarchical inverse dynamics to stabilize torque-controlled humanoid robots with different amount of actuation in ankle. In \cite{shafiee2016push}, the CoP modulation and angular momentum regulation have been used inside MPC based on the DCM dynamics to recover the robot from pushes in standing posture, while step adjustment has been added to this formulation to generate reactive walking patterns during walking \cite{shafiee2017robust}. In \cite{khadiv2016step}, a walking controller based on step location and timing adjustment inside a small-size Quadratic Program (QP) has been proposed which is a general approach for controlling robots with or without ankle actuation or even a point contact biped. This approach needs only one step horizon for guaranteeing the viability of the motion \cite{khadiv2017robust}. Furthermore, \cite{griffin2017walking} proposed a reactive walking pattern generator which uses a combination of step location and timing adjustment based on an analytical solution of the LIPM.

In this paper, we aim at using the approach in \cite{khadiv2016step} which automatically adapts step location and timing based on DCM measurement, and improving its robustness by adding CoP modulation and angular momentum regulation. The resulting walking pattern generator adapts step location and timing in the first stage, and in the second stage regenerates the DCM and angular momentum trajectories consistent with the first stage. In section II, after briefly outlining the formulation in \cite{khadiv2016step}, we present our proposed approach. Section III presents the simulation results and discussion. In section IV, we conclude the findings.\\

\section{{\Large W}ALKING {\Large P}ATTERN {\Large G}ENERATOR}
\begin{figure}[h]
	\centering
	\includegraphics[scale=0.59, trim ={4.2cm 14.5cm 2.0cm 2cm},clip]{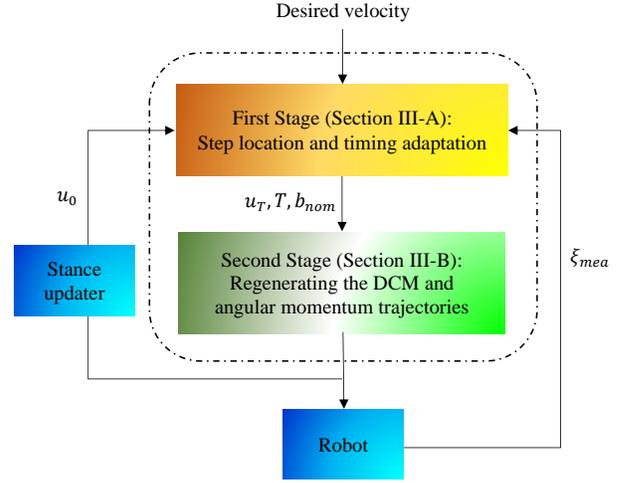}\par
	\caption{The proposed method framework. In the first stage, the location and landing time of the swing foot are adapted through a small-size Quadratic Program (QP) optimization, using DCM measurement. Then, the second stage regenerates the DCM and angular momentum trajectories at each control cycle, constraining the DCM at the end of the step to track the desired DCM.}
	\vspace{-1.5em}
	\label{diagram}
\end{figure}

The block diagram of the proposed walking pattern generator is shown in Fig. \ref{diagram}. In the first stage of the proposed method, based on the LIPM dynamics with point contact, the landing location and time of the swing foot are adapted at each control cycle using DCM measurement. The objective in the second stage is to achieve a desired DCM at the end of the step, based on adapted gait variables from the first stage. Note that in the second stage, the LIPM with flywheel is used to generate dynamically consistent DCM and angular momentum trajectories. 

\subsection{First stage: Step location and timing adaptation \cite{khadiv2016step}}
In this stage, the nominal values of the gait variables are determined based on a robustness criteria to comply with the desired walking velocity as well as to satisfy the dynamic and kinematic constraints \cite{khadiv2016step}. The nominal values of the step length ($ L_{nom} $), step width ($ W_{nom} $), and the step time ($ T_{nom} $) are then used to construct an optimization problem which adapts the gait variables as close as possible to the nominal values based on the current measurement of the DCM.\\
The LIPM (with point contact) solution in terms of the next step location and duration, and the DCM offset can be written down as \cite{khadiv2016step}:
\begin{equation}
{{u}_{T}} = 
({{\xi }_{cur}} - {{u}_{0}})
{{e}^{{{\omega }_{0}}(T-t)}} + 
{{u}_{0}} - b \quad,\quad 0\le t\le T
\label{xi_T}
\end{equation}

in which $ {\xi_{cur}} $ and $ u_0 $ are the current DCM and step location, respectively. $ b $ is the DCM offset (the distance between the DCM at the end of step and next step location) and $ \omega_0 $ is the natural frequency of the pendulum (${{\omega}_{0}}=\sqrt{{g}/{{h}}\;}$, where g is the gravity constant, and $h$ is the CoM height). In order to convert the nonlinear equation 
\eqref{xi_T} to linear form, a transformation could be utilized as \cite{khadiv2016step}:
\begin{equation}
{\tau}=
{{e}^{{{\omega}_{0}}(T)}}
\to 
T=
\dfrac{1}{w_0}
{\log{\tau}}
\end{equation}
This transformation makes the main constraint of the problem linear with respect to $ \tau $. Hence, the following QP is solved for the landing location and time of the swing foot as well as the DCM offset:
\begin{equation}
\begin{split}
\underset{{{u}_{T}},\tau ,b}{\mindot} \; & 
{{\alpha }_{1}}
{\norm{{u}_{T}-
		\begin{bmatrix}
		{L}_{nom} \\ 
		{W}_{nom} \\ 
		\end{bmatrix}
}}^{2}+
{{\alpha }_{2}}
{\abs
	{\tau -{{\tau }_{nom}}}
	^{2}}\\
& +{{\alpha }_{3}}
{\norm{b-
		\begin{bmatrix}
		{b}_{x,nom} \\ 
		{b}_{y,nom} \\
		\end{bmatrix}}
}^{2}\\
& \text{s.t.}
\begin{bmatrix}
{{L}_{min}}  \\
{{W}_{min}} \\
\end{bmatrix}
\le {{u}_{T}} \le
\begin{bmatrix}
{{L}_{max}}  \\
{{W}_{max}}  \\
\end{bmatrix} \\ 
& {{u}_{T}}+b=({{\xi }_{mea}}-{{u}_{0}}){{e}^{-{{\omega }_{0}}t}}\tau +{{u}_{0}} \\
& {{e}^{{{\omega }_{0}}{{T}_{min}}}}\le \tau \le {{e}^{{{\omega }_{0}}Tmax}}
\end{split}
\label{fi_QP}
\end{equation}
The adapted swing foot landing location and time are realized using the swing foot trajectory generation method in \cite{khadiv2016step}.\\

\subsection{Second stage: Regenerating the DCM and angular momentum trajectories}
\begin{figure}
	\centering
	\includegraphics[scale=0.61, trim ={2.9cm 16.cm 10.3cm 4.35cm},clip]{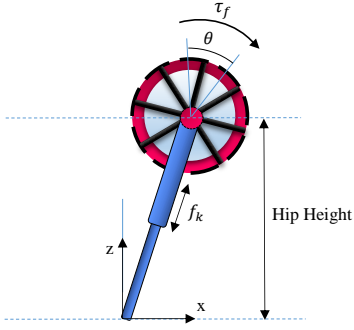}\par
	\caption{LIPM with flywheel}
	\vspace{-1.5em}
	\label{flywheel}
\end{figure}

In the previous subsection, we outlined a reactive stepping planner to regenerate the next step location and time at each control cycle. In this subsection, we use the LIPM with flywheel \cite{pratt2006capture} to generate the DCM and trunk orientation trajectories (as an estimation of the whole-body angular momentum around the CoM). In the LIPM with flywheel, decoupled linear and angular momentum are considered, where a limited amount of angular momentum can be generated around the CoM. The dynamics of this system can be written down as:
\begin{equation}
\ddot{x} = 
{{{\omega}}_{0}}^{2}(x-{{x}_{CMP}})
\end{equation}
in which $ x $ is a 2-D vector containing CoM horizontal components (the vertical component has a fixed value $h$). $x_{CMP}$ is the vector of Central Moment Pivot (CMP) location ($x_{CMP}={[{CMP}_x, {CMP}_y]}^T$) and can be specified as:
\begin{equation}
{{x}_{CMP}} = 
z+
\frac
{{{\tau }_{f}}}{mg}
\label{cmp}
\end{equation}
where $z$ stands for the ZMP (CoP) location and $ {\tau }_{f} $ is the amount of torque applied around the CoM (see Fig. \ref{flywheel}). It is clear in this equation that in the absence of $ {\tau }_{f} $, the ZMP and CMP coincide, and as a result the dynamics equations of the LIPM with flywheel are identical with the LIPM dynamics equations.
By considering the CoM ($x$) and DCM ($\xi=x+{\dot{x}}/{{\omega}_0}$) as the state variables, the LIPM dynamics in the state space form may be specified as:
\begin{subequations}
	\begin{align}
	\dot{x} &  = 	{{\omega }_{0}}(\xi -x)\\
	\dot{\xi } &  =  {{\omega }_{0}}(\xi -{{x}_{CMP}})
	\label{DCM}
	\end{align}
	\label{LIPM_dynamics}
\end{subequations}
Equation \eqref{LIPM_dynamics} decomposes the LIPM dynamics into its stable and unstable parts, where the CoM converges to the DCM and the DCM is pushed away by the CMP.\\
The goal here is to bring the DCM to the desired position which realizes the nominal DCM offset with respect to the updated step location:
\begin{equation}
{{\xi}_{des}}=
{{u}_{T}}+{b}_{nom}
\label{xi_des}
\end{equation}

in which $ {u}_{T} $ is obtained from QP
\eqref{fi_QP} and $ {b}_{nom} $ is the nominal DCM offset computed for having a desired walking velocity \cite{khadiv2016step}.\\
By considering the dynamics equation of the LIPM with flywheel and replacing 
\eqref{cmp} into
\eqref{DCM}, we obtain:
\begin{equation}
\dot{\xi }   =  {{\omega }_{0}}(\xi - z -
\frac
{{{\tau }_{f}}}{mg})
\label{replacing}
\end{equation}
According to Fig. \ref{flywheel}, the angular momentum dynamics around the CoM can be specified as below:
\begin{equation}
{\tau_{f}}=j{\ddot{\theta}}
\label{tau_f}
\end{equation}
Where $ j $ is the trunk inertia. In fact, in this equation we only considered the contribution of the upper body (trunk) angular momentum around the CoM and neglected the effect of other rotating parts. Therefore, by substitution of 
\eqref{tau_f} into
\eqref{replacing} and some mathematical manipulations, the dynamics equation can be formulated as:
\begin{equation}
z=\xi -\frac{1}{{{\omega }_{0}}}\dot{\xi }-\frac{j}{mg}\ddot{\theta}
\label{horizontal_z}
\end{equation}
where $ \xi $ is the horizontal position of the DCM, $ {\dot{\xi}} $ its horizontal velocity and $ \ddot{\theta} $ is the rate of angular momentum (angular acceleration of the trunk). This approximation naturally decouples the forward and lateral motions of the robot. In order to discretize the dynamics of the system, we assume the upper body jerk and the DCM acceleration to be constant over each sampling interval $ T $, hence at times $ t_k = kT $ with $ k = 1, 2, \ldots $ we have:
\begin{equation}
\begin{split}
&\ddot{\xi }_{k}={{\ddot{\xi }}_{k}}(kT) ;\enspace
{{\ddot{\xi }}_{k}}=\mathit{const.}\\
&\dddot{\theta}_k=\dddot{\theta}_k(kT) ;\enspace
\dddot{\theta}_k=\mathit{const.}\\
\end{split}
\end{equation}
Therefore, by considering the state variables vector as:
\begin{equation}
{{\hat{x}}_{k}}=
\begin{bmatrix}
{{\xi}(t_{k})}  \\
{{{\dot{\xi}}}(t_{k})}  \\
{{\theta}(t_{k})}  \\
{{{\dot{\theta}}}(t_{k})}  \\
{{{\ddot{\theta}}}(t_{k})}  \\
\end{bmatrix}
\label{statevariables}
\end{equation}
and focusing on the decoupled motion in the sagittal direction, the equation
\eqref{horizontal_z} leads to:
\begin{equation}
z_{k}^{x}=
\begin{bmatrix}
1 & 
-\frac{1}{{{\omega }_{0}}} & 
0 & 
0 & 
-\frac{j}{mg}  \\
\end{bmatrix}
{{\hat{x}}_{k}}
\label{z_x}
\end{equation}
Now, we can compute the corresponding discrete dynamics as:
\begin{equation}
{{\hat{x}}_{k+1}}=A{{\hat{x}}_{k}}+B{{u}_{k}}
\enspace,\enspace
{{u}_{k}}=
{\begin{bmatrix}
	{{{\ddot{\xi }}}_{k}} 
	\enspace 
	{\dddot{\theta}_{k}}  
	\end{bmatrix}}^T
\label{discerete_dynamic}
\end{equation}
where
\begin{equation}
A=
\begin{bmatrix}
1 & T & 0 & 0 & 0  \\
0 & 1 & 0 & 0 & 0  \\
0 & 0 & 1 & T & {{T}^{2}}/2  \\
0 & 0 & 0 & 1 & T  \\
0 & 0 & 0 & 0 & 1  \\
\end{bmatrix}
\enspace,\enspace
B=\begin{bmatrix}
{{T}^{2}}/2 & 0  \\
T & 0  \\
0 & {{T}^{3}}/6  \\
0 & {{T}^{2}}/2  \\
0 & T  \\
\end{bmatrix}
\end{equation}

Based on the discrete-time state space model of the system, we present a constrained predictive controller which optimizes the future state variables and control inputs to bring the predicted output as close as possible to the desired values. In fact, at this stage, our main goal is to realize the desired DCM at the end of the current step consistent with the desired DCM offset and adapted step location and time. Hence, the resulting optimization problem regenerates the DCM and angular momentum trajectories at each control cycle employing the upper body angular jerk and the DCM acceleration. To achieve this purpose, we denote the future state variables as:
\begin{equation*}
{\boldsymbol{\upxi }_{k+1}}=
\begin{bmatrix}
{{\xi }_{k+1}}  \\
\vdots\\
{{\xi }_{k+N}}  \\
\end{bmatrix},
{\boldsymbol{\dot{\upxi }}_{k+1}}=
\begin{bmatrix}
{{{\dot{\xi }}}_{k+1}}  \\
\vdots\\
{{{\dot{\xi }}}_{k+N}}  \\
\end{bmatrix},
{\boldsymbol{\uptheta }_{k+1}}=
\begin{bmatrix}
{{\theta }_{k+1}}  \\
\vdots\\
{{\theta }_{k+N}}  \\
\end{bmatrix}
\end{equation*}
\begin{equation}
{\boldsymbol{\dot{\uptheta }}_{k+1}}=
\begin{bmatrix}
{{{\dot{\theta }}}_{k+1}}  \\
\vdots\\
{{{\dot{\theta }}}_{k+N}}  \\
\end{bmatrix},
{\boldsymbol{\ddot{\uptheta }}_{k+1}}=
\begin{bmatrix}
{{{\ddot{\theta }}}_{k+1}}  \\
\vdots\\
{{\ddot{\theta }}_{k+N}}  \\
\end{bmatrix}
\end{equation}
in which $ N $ is the prediction horizon. We can compute the augmented dynamic model of the system recursively using
\eqref{z_x} and
\eqref{discerete_dynamic} as follows:
\begin{equation}
\begin{aligned}
& {\boldsymbol{\upxi }_{k+1}}=
{{P}_{1s}}
{{{\hat{x}}}_{k}}+
{{P}_{1u}}
{\boldsymbol{\mathrm{U}}_{k}} \\ 
& {{\boldsymbol{\dot{\upxi }}}_{k+1}}=
{{P}_{2s}}
{{{\hat{x}}}_{k}}+
{{P}_{2u}}
{\boldsymbol{\mathrm{U}}_{k}} \\ 
& {\boldsymbol{\uptheta }_{k+1}}=
{{P}_{3s}}
{{{\hat{x}}}_{k}}+
{{P}_{3u}}
{\boldsymbol{\mathrm{U}}_{k}} \\ 
& {\boldsymbol{\dot{\uptheta }}_{k+1}}=
{{P}_{4s}}
{{{\hat{x}}}_{k}}+
{{P}_{4u}}
{\boldsymbol{\mathrm{U}}_{k}} \\ 
& {\boldsymbol{\ddot{\uptheta }}_{k+1}}=
{{P}_{5s}}
{{{\hat{x}}}_{k}}+
{{P}_{5u}}
{\boldsymbol{\mathrm{U}}_{k}} \\ 
& {{Z}_{k+1}}=
{{P}_{zs}}
{{\hat{x}}_{k}}+
{{P}_{zu}}
{\boldsymbol{\mathrm{U}}_{k}}
\end{aligned}
\end{equation}
where
\begin{equation}
{\boldsymbol{\mathrm{U}}_{k}}=
{\begin{bmatrix}
	{{{\ddot{\xi}}}_{k}},  
	\ldots,
	{{{\ddot{\xi}}}_{k+N-1}},  
	{{\dddot{\theta}}_{k}},  
	\ldots,
	{\dddot{\theta}}_{k+N-1},  
	\end{bmatrix}}^{T}
\end{equation}

Now, we define the optimization problem as follows:
\begin{equation}
\begin{aligned}
\min. \;
&{{\beta}_{1}}\norm{{{Z}_{k+1}} - {{Z}_{k+1}^{ref}}} + 
{{\beta}_{2}}\norm{{\ddot{\xi}_k}} + 
{{\beta}_{3}}\norm{{\dddot{\theta}_k}} \\
&+{{\beta}_{4}}\norm{{\dot{\theta}_{k+1}}} + 
{{\beta}_{5}}\norm{{{\xi}_{T}} - {{\xi}_{des}}}\\
\text{s.t.}\quad &
\abs{{\ddot{\theta }}}
\le
\ddot{\theta }_\text{max} \\
&\theta_{min}
\le \theta \le
\theta_{max} \\
&Z_{k+1} \in support \, polygon \\
\end{aligned}
\label{fii_QP}
\end{equation}

where $ {Z}_{k+1}^{ref} $ is the future reference ZMP (CoP) and considered to be in the middle of the support polygon in order to distance the ZMP from edges of the support polygon. Furthermore, $\xi_T$ is the DCM at the end of the step. We give $\beta_5$ a large value compared to the other weights to enforce the optimizer to bring the DCM to its desired value at the end of the step. Note that in this algorithm the matrices $ P_{1s}, P_{2s}, \ldots, P_{zs} $ and $ P_{1u}, P_{2u}, \ldots , P_{zu} $ should be recomputed at each control cycle, because the prediction horizon may be changed being proportional to the adapted step time.
We imposed constraints on the trunk angle and angular acceleration \eqref{fii_QP} to take into account the robot physical limitations in applying angular momentum around the CoM.\\
\vspace{-1.0em}
\section{{\Large R}ESULTS AND {\Large D}ISCUSSIONS}
In this section, we present two simulation scenarios to show the effectiveness of our proposed algorithm. In the first scenario, we present the results obtained from simulating the LIPM with flywheel. In this scenario, we push the robot and show the robustness of the gaits in the presence of disturbances, where our controller with angular momentum regulation and CoP modulation is used. In the second scenario, we compare our approach to \cite{khadiv2016step} in terms of robustness against pushes.\\
\vspace{-1.0em}
\subsection{Simulation results using the LIPM with flywheel}
In the first scenario, we simulate motion of the LIPM with flywheel abstraction of a biped robot controlled by our proposed approach. The physical properties of the model and the robot constraints are given in TABLE I. Also, in Table II, weighting coefficients and physical characteristics of the robot are given. In this scenario, the resulting convex hull of the set of contact points between the feet and the ground creates the support polygon in each single or double support phases. The feasible area for step location is computed with respect to the current state of the stance foot.
\begin{table}[h!]
	\centering
	\captionof{table}{Physical properties of the abstract model}
	\begin{tabular}{llll}
		\hline
		Variable      &  Description            &  min        &  max  \\  \hline
		$L$      &  Step length              &  -50 ($cm$)        &     50 ($cm$) \\  
		$W_{right}$      & Step width (right)             &  -10 ($cm$)        &20 ($cm$)      \\  
		$W_{left}$  &  Step  width (left) & -20 ($cm$)      &  10 ($cm$)  \\
		$T$   &  Step duration&  0.3 ($sec$)        &  1.0 ($cm$)  \\  \hline
	\end{tabular}
\end{table}

\begin{table}[h!]
	\centering
	\captionof{table}{Anthropomorphic proportions of the robot}
	\begin{tabular}{llll}
		\hline
		Variable                   &  Description         &  Value           &      \\  \hline
		$m$                        &  Body mass           &  60 kg           &      \\  
		$h $                       &  CoM height          &  0.8 m           &      \\ 
		$\psi_{max}$ (forward)     &  Max trunk rotation  &  $\pi/3$ rad     &      \\  
		$\psi_{min}$ (backward)    &  Min trunk rotation  &  $-\pi/3$ rad    &      \\  
		$j$                        &  Trunk inertia       &  8 kg.m$^2$      &      \\  
		$\tau_{max}$               &  Max torque on upper-body      &  150 N.m         &      \\  
		$\alpha_1$                 &  Control weight      &  1 $m^{-1}$    &      \\  
		$\alpha_2$                 &  Control weight      &  1 $N^{-1}.m^{-1}$    &      \\
		$\alpha_3$                 &  Control weight      &  1000 $m^{-1}$ &      \\
		$\beta_1$                  &  Control weight      &  1 $m^{-1}$    &      \\
		$\beta_2$                  &  Control weight      &  1 $m^{-1}.s^2$    &      \\
		$\beta_3$                  &  Control weight      &  1 $rad^{-1}.s^3$    &      \\
		$\beta_4$                  &  Control weight      &  1 $rad^{-1}.s$    &      \\  
		$\beta_5$                  &  Control weight      &  100000 $m^{-1}$    &      \\  \hline
		\vspace{-2.3em}
	\end{tabular}
\end{table}

 \begin{figure}[h!]
	\centering
	\includegraphics[scale=0.395, trim ={.1cm 8.5cm .2cm 8.7cm},clip]{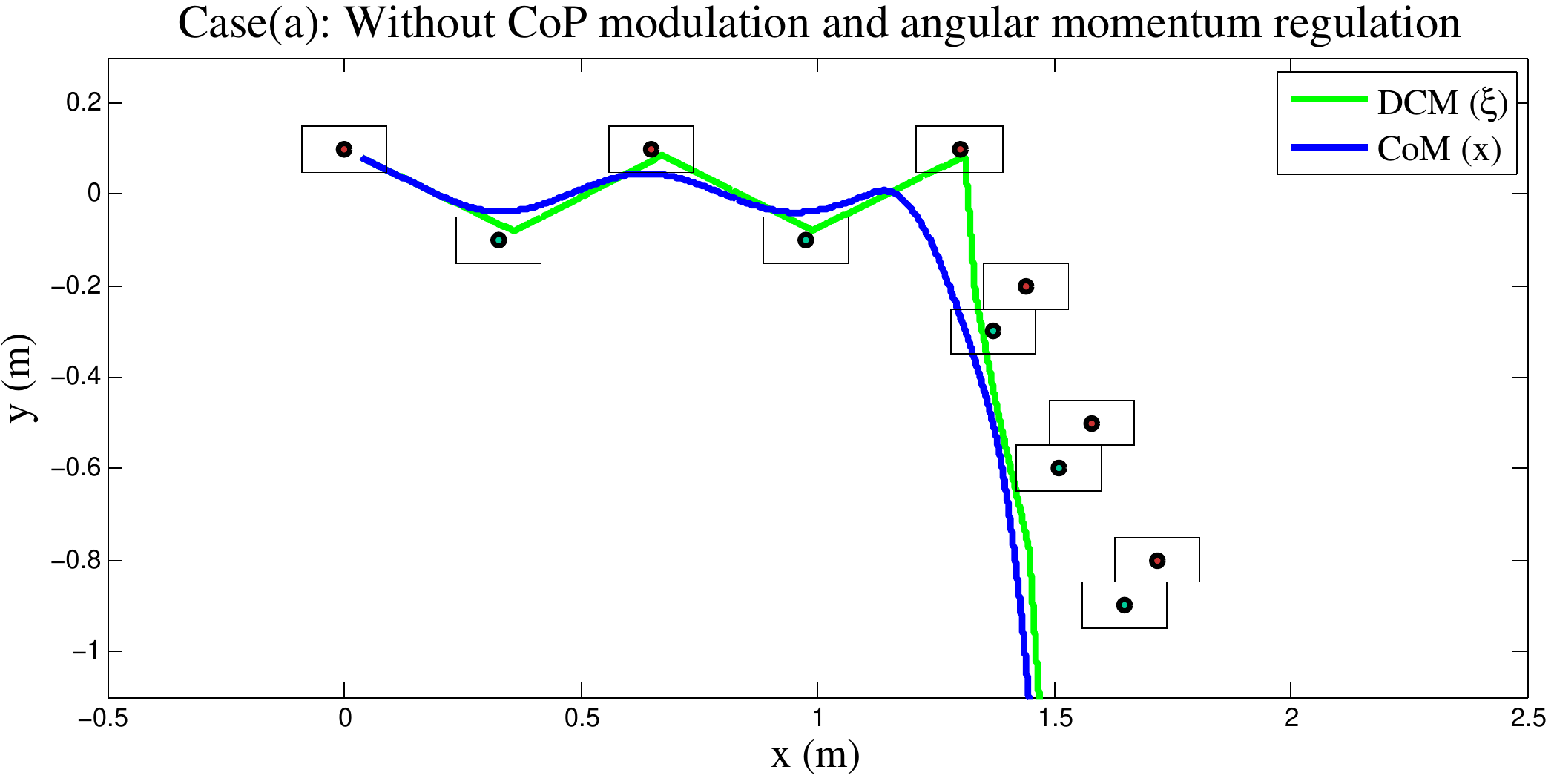}\par
	\label{fig_3a}
	\centering
	\includegraphics[scale=0.405, trim ={.5cm 8.5cm .5cm 8.5cm},clip]{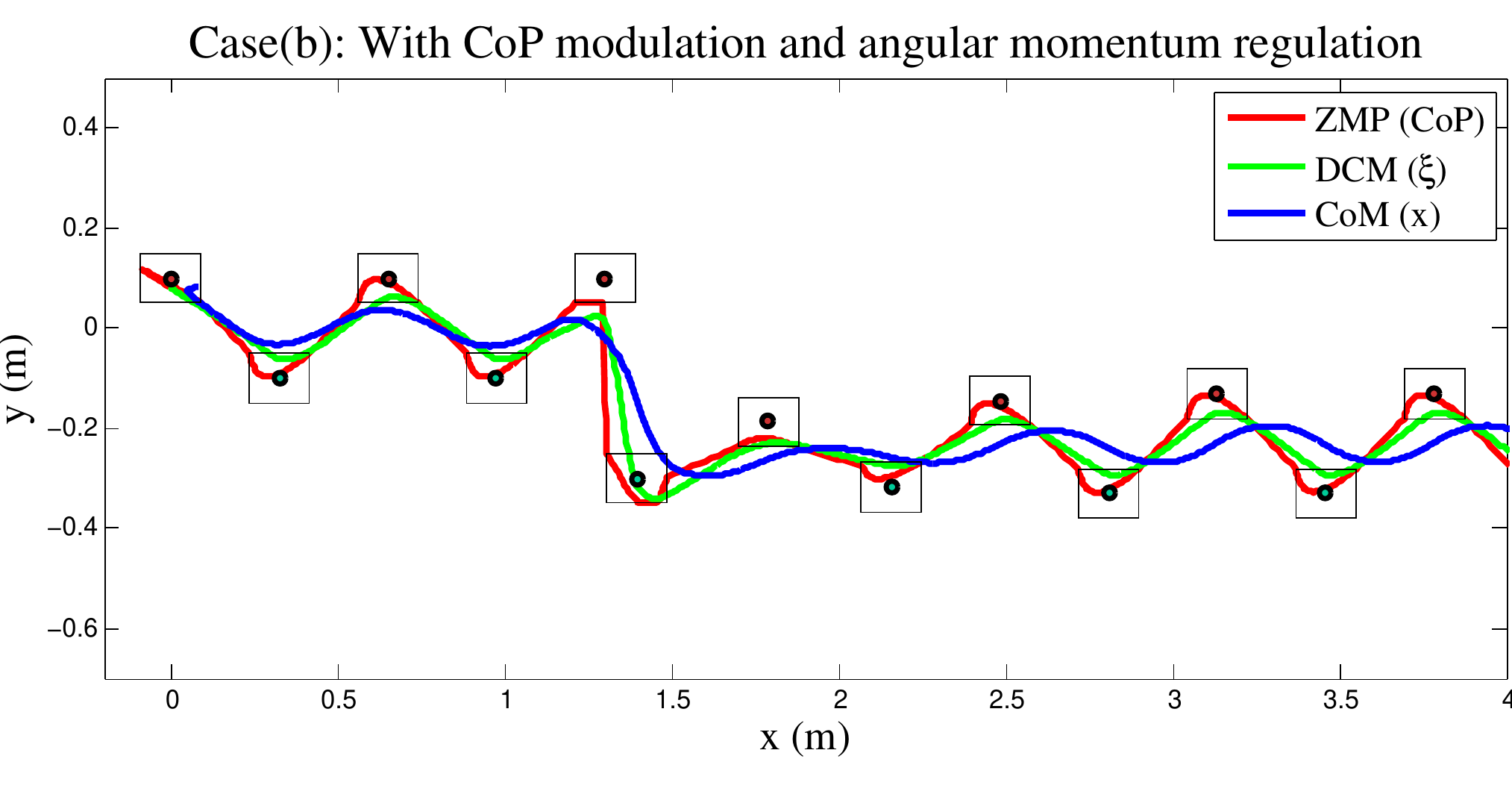}\par
	\caption{Comparing the Cartesian paths for the cases with and without CoP modulation and angular momentum regulation.} 
	\vspace{-0.5em}
	\label{fig_3b}
\end{figure}

  \begin{figure}[h!]
 	\centering
 	\includegraphics[scale=0.41, trim ={.2cm 8.2cm .3cm 8cm},clip]{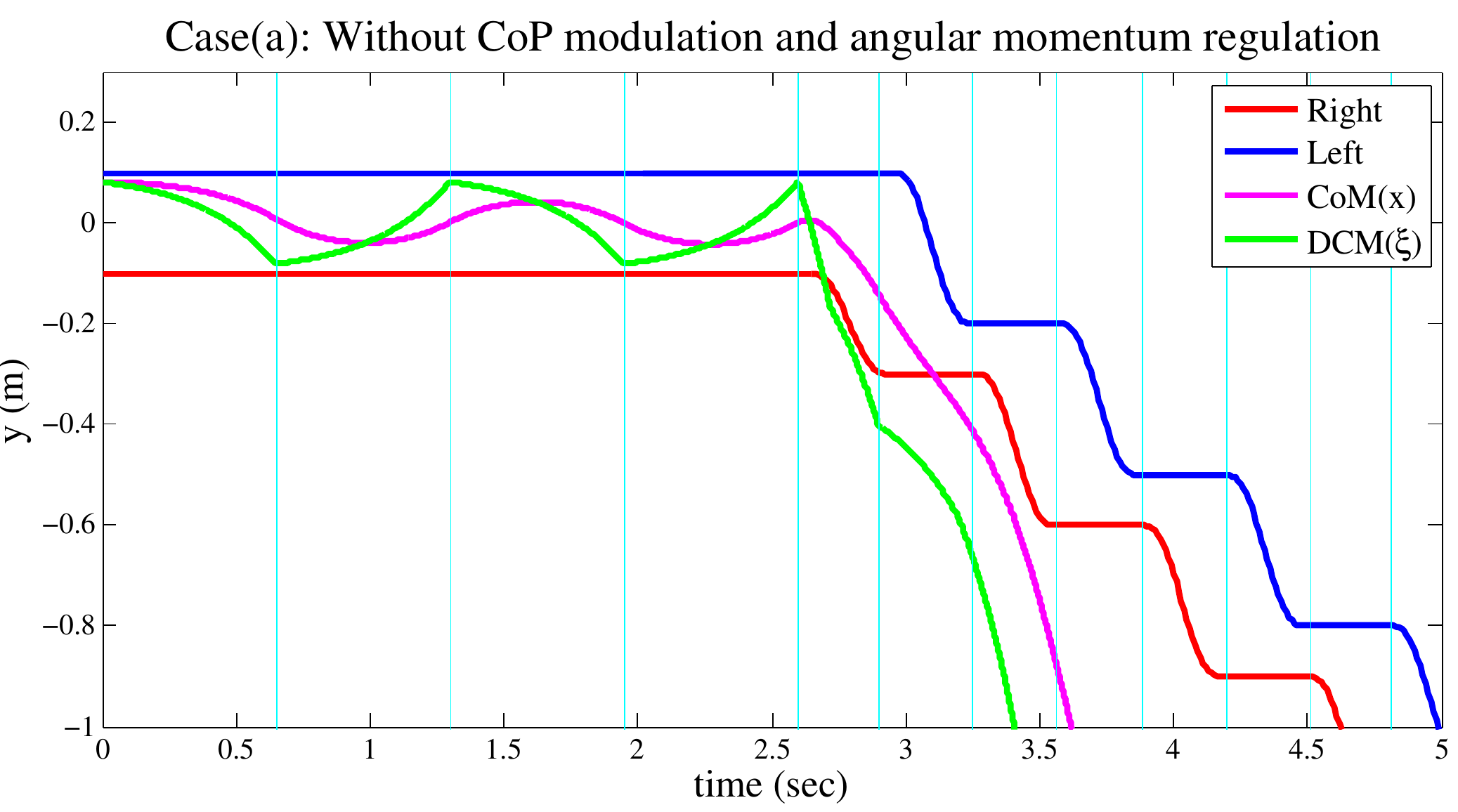}\par
 	\label{fig_4a}
 	\centering
 	\includegraphics[scale=0.41, trim ={.2cm 9cm 0.2cm 8.4cm},clip]{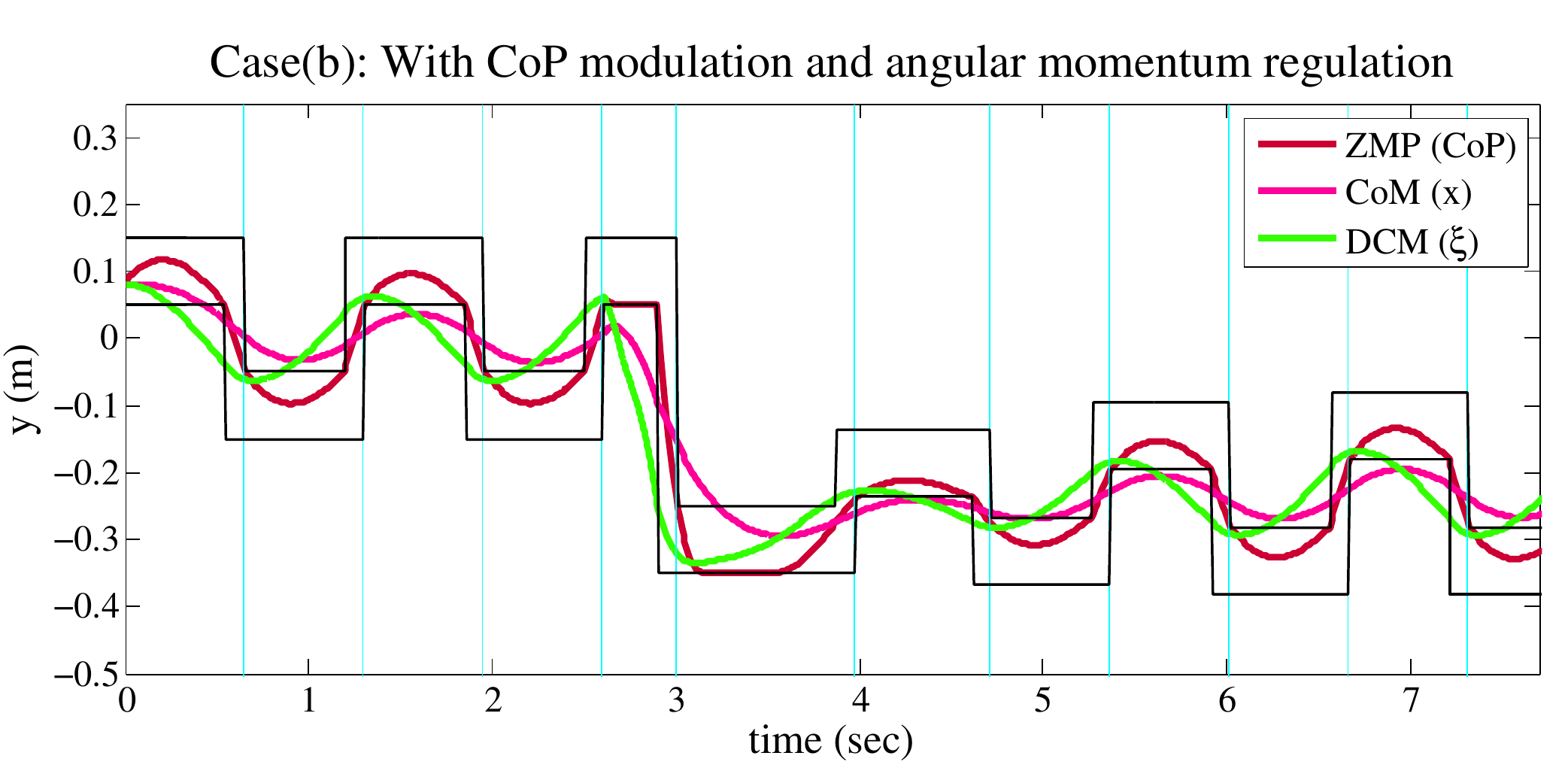}\par
 	\centering
 	\includegraphics[scale=0.45, trim ={1.6cm 11.5cm 0.2cm 9.6cm},clip]{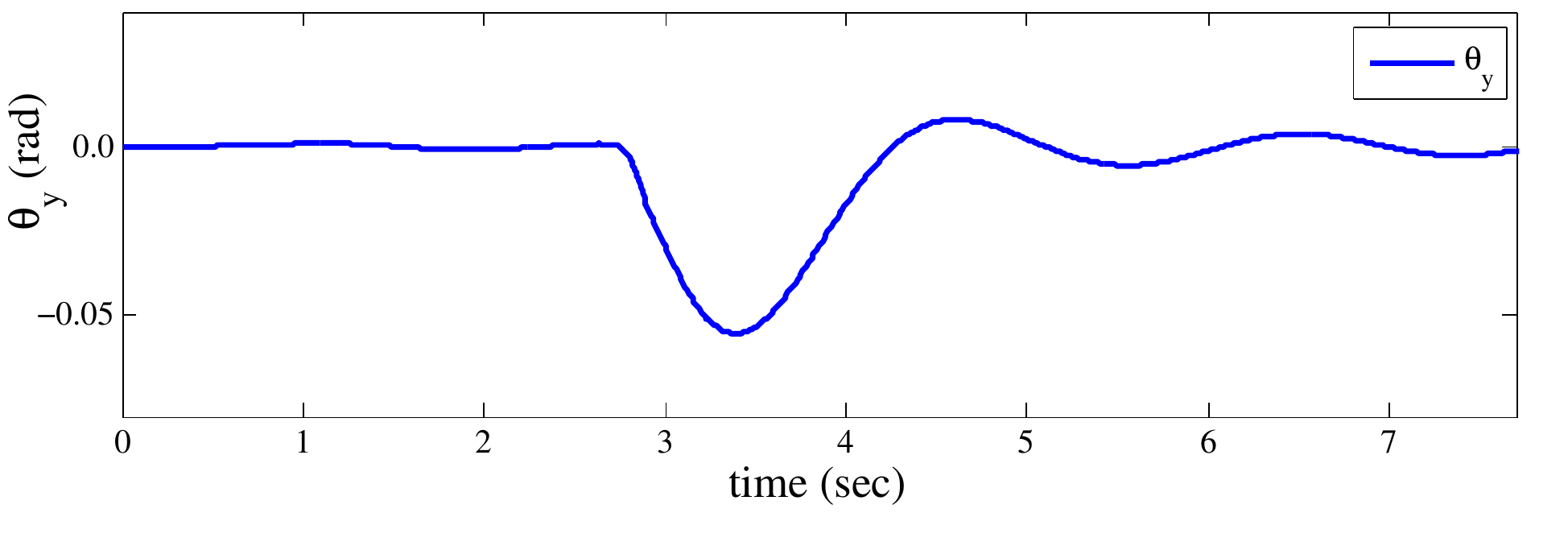}\par
 	\caption{Comparing the trajectories for the cases with and without CoP  modulation and angular momentum regulation. The vertical lines show the step duration.} 
 	\vspace{-0.5em}
 	\label{fig_4b}
 \end{figure}
 
We applied our proposed controller as described in section II. In this scenario, a velocity command $ (v_x=0.5 m/s) $ for forward walking is given. Based on the limitations specified in TABLE I, the nominal step length and step duration are computed as in \cite{khadiv2016step}, and the desired DCM offset is computed using \eqref{xi_des}. After four steps, the robot is pushed at $ t=2.6 $ s to the right direction with a force $ F=315 N $, during $ \Delta t = 0.1 s $. We conduct two simulations to compare the results of the cases with and without angular momentum regulation and CoP modulation.\\
 Fig. \ref{fig_3b} shows an example gait in the Cartesian space resulted from this simulation. For the case (a) illustrated in Fig. \ref{fig_3b}, we employ \eqref{fi_QP} for step length and timing adjustment, using DCM measurement. In this simulation, the LIPM (point-contact) is used to compute the next step location and duration and the DCM offset by minimizing the error between the gait variables and their nominal values. As it can be observed, the robot cannot recover from the push using only step location and timing adjustment and the DCM diverges. 

In the case (b) in Fig. \ref{fig_3b}, we solve the optimization procedure in \eqref{fii_QP} after \eqref{fi_QP} at each control cycle to regenerate the DCM and angular momentum trajectories.
 In this case, for the first four steps where there is no disturbance, all the nominal values are realized. After pushing the robot, the controller employs a combination of step location and timing adjustment, as well as angular momentum regulation and CoP modulation to recover the robot from the push. Obtained results show that after this severe push, the CoP leans towards the inner edge of the support polygon (Fig. \ref{fig_3b} (b)), and the upper body of the robot sways to avoid falling (Fig. \ref{fig_4b} (b)). As it can be observed in Fig. \ref{fig_3b} (b), after rejecting the push, the robot resumes its stepping with the desired velocity in forward direction and upright upper body. Furthermore, we can see that our approach can tolerate more severe pushes compared to the case without angular momentum regulation and CoP modulation.\\
  \begin{figure}[h!]
	\centering
	\includegraphics[scale=0.43, trim ={1.5cm 8.8cm 1cm 10cm},clip]{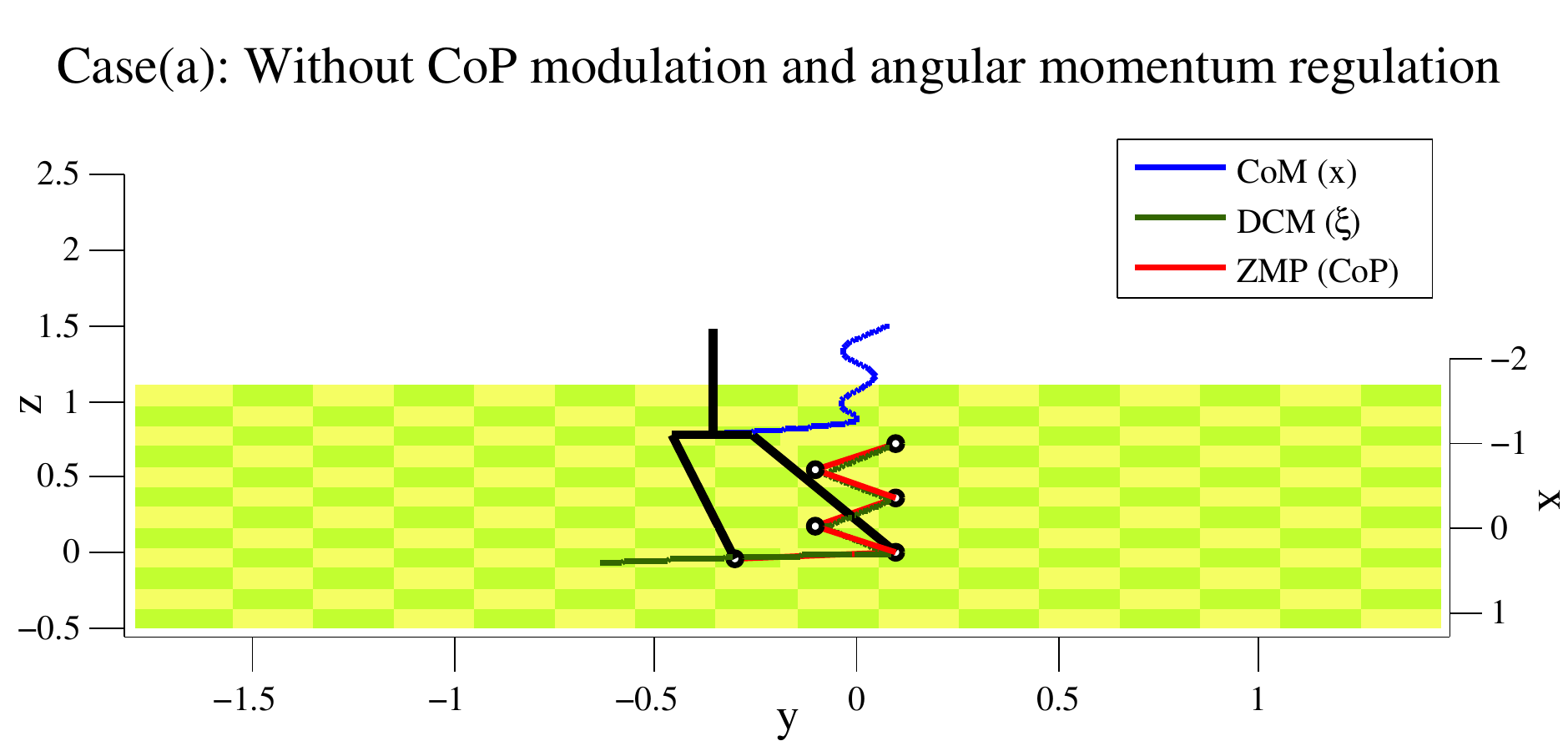}\par
	\vspace{-1.5em}
	\label{fignew_a}
\end{figure}
\begin{figure}[h!]
	\centering
	\includegraphics[scale=0.5, trim ={2.0cm 7.5cm 2cm 9.5cm},clip]{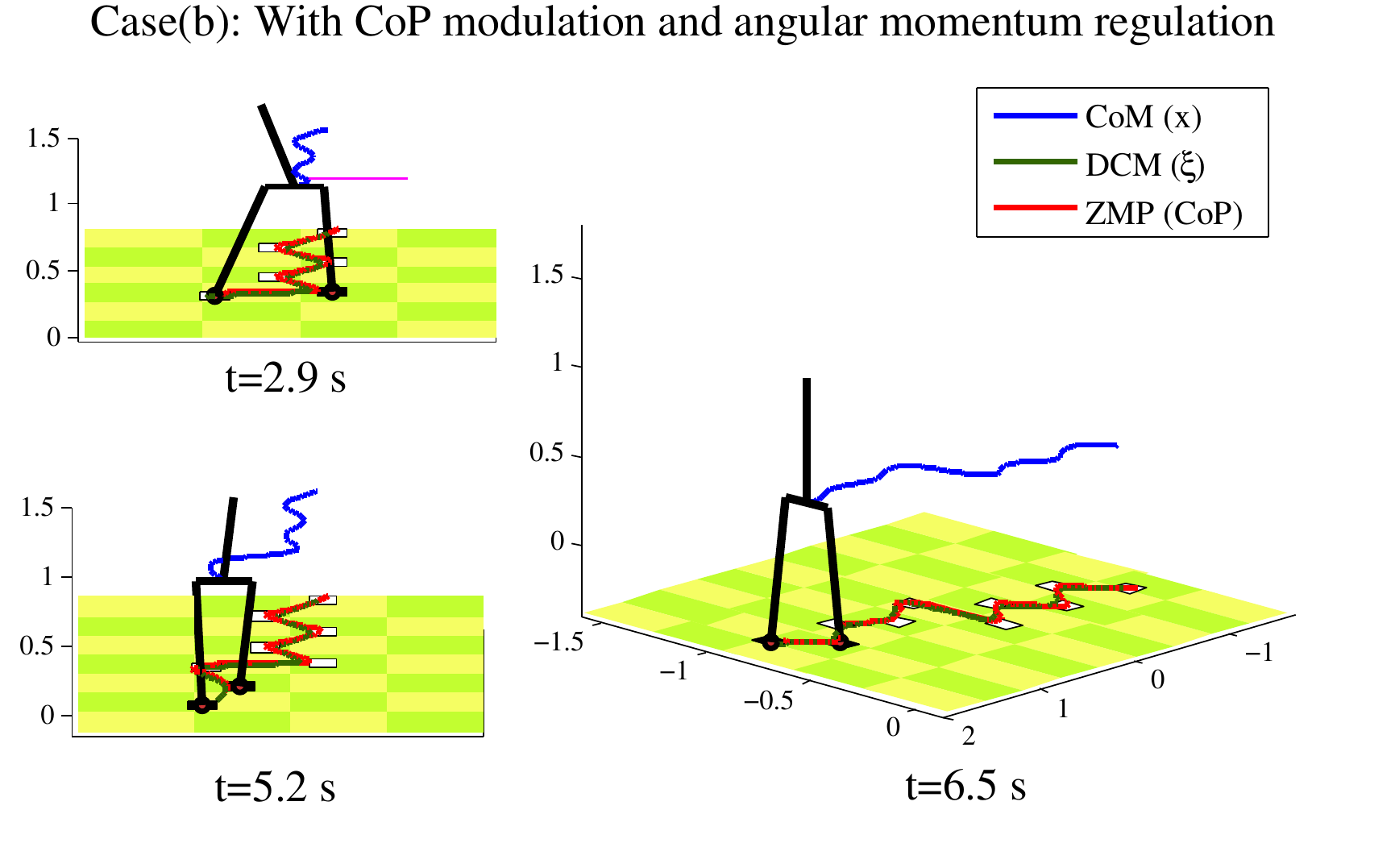}\par
	\caption{Cartooned motion of the robot. Case (a): without CoP modulation and angular momentum regulation. Case (b): with CoP modulation and angular momentum regulation}
	\label{fignew_b}
\end{figure}
Fig. \ref{fig_4b} illustrates the time history of the DCM, CoM and the feet in lateral direction during this simulation scenario for both cases. The vertical lines (Cyan) show the duration of each step. We can see in this figure that both cases adapt step timing together with step location, when the disturbance is exerted. In the case (a), the DCM diverges since adapting only the step location and timing is not enough to keep the robot states from diverging. However, in the case (b), angular momentum regulation and CoP modulation are also adopted to recover the robot from the disturbance. The robot motions for both cases are cartooned in Fig. \ref{fignew_b}.\\

\subsection{Comparison with \cite{khadiv2016step}}
\begin{figure}[h!]
	\centering
	\includegraphics[scale=0.45, trim ={1.4cm 8.6cm .2cm 7.7cm},clip]{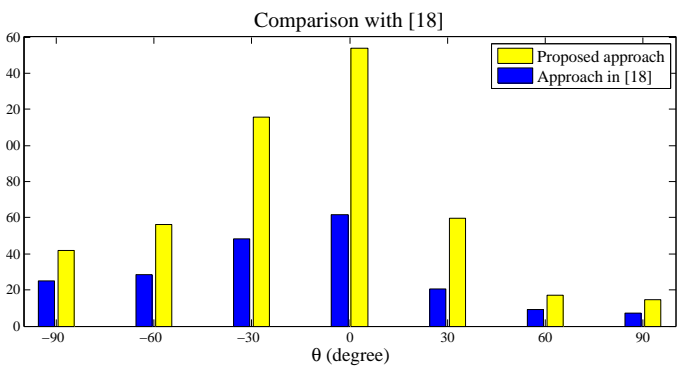}\par
	\caption{comparison with the approach in \cite{khadiv2016step}. $ \psi $ is the angle between the direction of motion and the push direction. For each simulation, a force during $ \Delta t =0.1 s $ is applied at the start of a step in which the left foot is stance.}
	\vspace{-1.5em}
	\label{fig7}
\end{figure}

In the second scenario, we compare the robustness of our proposed controller with the proposed approach in \cite{khadiv2016step}. We do this to perceive the effect of CoP modulation and angular momentum regulation on the gait robustness of biped robots. The optimizer in \cite{khadiv2016step} reactively generates the footstep location and duration in real-time, but the CoP is considered to be fixed and the the angular momentum is not regulated.\\
We applied the same parameters for both approaches and computed the maximum push that each approach can recover from in various directions (Fig. \ref{fig7}). The value $ \psi $  is the angle between the direction of motion and the push direction (counterclockwise). In order to gain symmetric results, we consider stepping with zero velocity, while similar results can be obtained for other desired velocities.
For each simulation, a force during $ \Delta t =0.1 s $ is applied at the start of a step in which the left foot is stance. The nominal values of step time in single support and double support phases are 0.55 s and 0.1 s, respectively. It should also be noted that in \cite{khadiv2016step} bipedal gait does not have double support phases and $ T_{nom} $ is $ 0.65 s $.\\
As it can be observed in Fig. \ref{fig7}, the presented approach in this work with CoP modulation and  angular momentum regulation can recover from more severe pushes compared to the approach in \cite{khadiv2016step}. We can also see in this figure that for the lateral direction, to avoid the feet collision, the amount of push that the controller can reject is less than the other direction. \\

\section{{\Large C}ONCLUSIONS}
In this paper, we proposed an approach to employ a combination of CoP manipulation, angular momentum regulation, as well as step location and timing adjustment to make biped walking gaits robust. The first stage of this approach employs the LIPM with point contact to adjust step location and timing based on DCM measurement inside a QP. The second stage exploits the LIPM with flywheel to simultaneously generate dynamically consistent trajectories for the DCM and angular momentum around the CoM.  In the proposed approach, after a disturbance, the step location and timing are adjusted and the DCM and angular momentum trajectories are consistently regenerated. Simulation results show that using angular momentum regulation and CoP modulation significantly increases the robustness capability of the existing walking planners.
\\

\bibliography{Master}
\bibliographystyle{IEEEtr}

\end{document}